\documentclass[runningheads]{llncs}
\usepackage{graphicx}

\usepackage{tikz}
\usepackage{comment}
\usepackage{amsmath,amssymb} 
\usepackage{color}

\usepackage[accsupp]{axessibility}  

\usepackage{booktabs}
\usepackage{siunitx}

\usepackage[para,online,flushleft]{threeparttable}
\usepackage{makecell}
\usepackage{booktabs}
\usepackage{multirow}
\usepackage{bm}
\usepackage{adjustbox}
\usepackage{pifont}
\usepackage{caption}
\usepackage{enumitem}
\usepackage{subcaption}

\usepackage[pagebackref,breaklinks,colorlinks]{hyperref}
\usepackage{xcolor}
\usepackage[normalem]{ulem}

\usepackage{floatrow}
\newfloatcommand{capbtabbox}{table}[][\FBwidth]

\usepackage[capitalize]{cleveref}
\crefname{section}{Sec.}{Secs.}
\Crefname{section}{Section}{Sections}
\Crefname{table}{Table}{Tables}
\crefname{table}{Tab.}{Tabs.}

\newcommand{\cmark}{\ding{51}}%
\newcommand{\xmark}{\ding{55}}%

\begin{document}

\pagestyle{headings}
\mainmatter
\def\ECCVSubNumber{4478}  

\title{Compiler-Aware Neural Architecture Search for On-Mobile Real-time Super-Resolution} 

\newcommand*\samethanks[1][\value{footnote}]{\footnotemark[#1]}
\titlerunning{Compiler-Aware NAS for On-Mobile Real-time SR}
%
\author{
Yushu Wu\thanks{Both authors contributed equally.}\inst{1} \and
Yifan Gong\samethanks[1]\inst{1} \and
Pu Zhao\inst{1} \and
Yanyu Li\inst{1} \and
Zheng Zhan\inst{1} \and
Wei Niu\inst{2} \and
Hao Tang\inst{3} \and
Minghai Qin\inst{1} \and
Bin Ren\inst{2} \and
Yanzhi Wang\inst{1}
}
\authorrunning{Y. Wu, Y. Gong et al.}
%
\institute{Northeastern University, Boston MA 02115, USA 
\email{\{wu.yushu,gong.yifa\}@northeastern.edu} \and
College of William and Mary, Williamsburg  VA 23185, USA \and 
CVL, ETH Zürich, Zürich 8092, Switzerland
}

\maketitle

\begin{abstract}
   Deep learning-based super-resolution (SR) has gained tremendous popularity in recent years because of its high image quality performance and wide application scenarios. However, prior methods typically suffer from large amounts of computations and huge power consumption, causing difficulties for  real-time inference, especially on resource-limited platforms such as mobile devices. To mitigate this, we propose a compiler-aware SR neural architecture search (NAS) framework that conducts depth search and per-layer width search with adaptive SR blocks. The inference speed is directly taken into the optimization  along with the SR loss to derive SR models with high image quality while satisfying the real-time inference requirement. Instead of measuring the speed on mobile devices at each iteration during the search process, a speed model incorporated with compiler optimizations is leveraged to predict the inference latency of the SR block with various width configurations for faster convergence. With the proposed framework, we achieve real-time SR inference for implementing 720p resolution with competitive SR performance (in terms of PSNR and SSIM)   
on GPU/DSP of mobile platforms (Samsung Galaxy S21). Codes are available at \href{https://github.com/wuyushuwys/compiler-aware-nas-sr}{\texttt{link}}. 
\end{abstract}

\section{Introduction}
\label{sec:intro}

As a classic 
vision task, single-image-super-resolution (SISR) restores the original high-resolution (HR) image based on a down-sampled low-resolution (LR) one. It can be applied in various applications, such as low-resolution media data enhancement or video/image upscaling for high  resolution  display panels. 
Various classic \cite{irani1991improving,freeman2002example,timofte2013anchored,timofte2014a+} and deep learning (DL)-based \cite{dong2014learning,dong2016accelerating,shi2016real,yu2018wide,liu2020fast} SR methods have been proposed in the past. Compared with classic interpolation algorithms to improve  image/video resolution, DL-based methods take  advantage of   learning  mappings from  LR to HR images from external datasets. Thus most   recent SR works emerge in the DL area. However, one   major limitation of existing DL-based SR methods is their high  computation and storage overhead to  achieve superior image quality, leading to difficulties to implement real-time SR inference even on powerful GPUs, not to mention resource limited edge devices. Due to the ever-increasing popularity of mobile devices and interactive on-mobile applications (such as live streaming),  it is essential to  derive lightweight SR models with both high image quality and low on-mobile inference latency.

There exist several works targeting at efficient SR models, including using upsampling operator at the end of a network \cite{dong2016accelerating,shi2016real}, adopting channel splitting \cite{hui2019lightweight}, using wider activation \cite{yu2018wide}, and  combining lightweight residual blocks with variants of group convolution \cite{liu2020fast}. Neural architecture search (NAS) is applied  to derive the optimal architecture in  many vision tasks. Latest works \cite{chu2019fast,song2020efficient,lee2020journey,chu2020multi} try to derive fast, lightweight, and accurate SR networks via NAS. However, their models 
are still too large to be implemented on mobile devices. Furthermore, these methods usually take the parameter numbers and computation counts (such as multiply-accumulate (MAC) operations) into the optimization for model efficiency, without considering the actual on-mobile implementation performance such as the inference latency. The actual mobile deployment of SR mobiles has rarely been investigated. The most relevant works are the winner of the PIRM challenge \cite{vu2018fast}, MobiSR \cite{lee2019mobisr}, and work \cite{Zhan2021AchievingOR}. But they either require nearly one second per frame for inference, far beyond real-time, or take a long search time. 

Targeting at achieving real-time inference of accurate SR model for 720p resolution on various resource-limited hardware such as mobile GPU and DSP, this paper proposes a compiler-aware NAS framework. An adaptive SR block is introduced to conduct the depth search and per-layer width search. Each convolution (CONV) layer is paired with a mask layer in the adaptive SR block for the width search, while the depth search is reached by choosing a path between the skip connection and the masked SR block. The mask can be trained along with the network parameters via gradient descent optimizers, significantly saving   training overhead. Instead of using MACs 
as the optimization target, the latency performance is directly incorporated into the objective function with the usage of a speed model. Our implementation can support real-time SR inference with competitive SR performance on various resource-limited platforms, including mobile GPU and DSP.
The contributions are summarized below:
\begin{itemize}[leftmargin=*]
    \item We propose a framework to search for the appropriate depth and per-layer width with  adaptive SR blocks. 
    \item We introduce a general compiler-aware speed model to predict the inference speed on the target device with corresponding compiler optimizations. 
    \item The proposed framework can directly optimize the inference latency, providing the foundations for achieving real-time SR inference on mobile.
    \item Our proposed framework can achieve real-time SR inference (with only tens of milliseconds per frame) for the implementation of 720p resolution with competitive SR performance (in terms of PSNR and SSIM) on mobile (Samsung Galaxy S21). Our achievements can facilitate various practical SR applications with real-time requirements such as live streaming or video communication. 
\end{itemize}

\section{Related Work}
\label{sec:related work}

\noindent \textbf{SR Works.} 
In recent years, most SR works have shifted their approaches from classic methods to DL-based methods with significant SR performance improvements. From the pioneering SRCNN \cite{dong2014learning} to later works with shortcut operator, dense connection, and attention mechanism \cite{kim2016accurate,lim2017enhanced,zhang2018residual,zhang2018image,dai2019second}, the up-scaling characteristic have dramatically boosted at the cost of high storage and computation overhead. Most of the works mentioned above even take seconds to process only one image on a powerful GPU, let alone mobile devices or video applications.

\noindent \textbf{Efficient SR.} 
Prior SR works are hard to be implemented on resource-limited platforms due to high computation and storage cost. To obtain more compact SR models, FSRCNN \cite{dong2016accelerating} postpones the position of the upsampling operator. IDN \cite{hui2018fast} and IMDN \cite{hui2019lightweight} utilize the channel splitting strategy.
CARN-M \cite{ahn2018fast} explores a lightweight SR model by combining efficient residual blocks with group convolutions. 
SMSR \cite{wang2021exploring}  learns sparse masks to prune
redundant computation for efficient inference. ASSLN \cite{zhang2021aligned} and  SRPN \cite{zhang2021learning}  leverage structure-regularized pruning and
impose regularization on the pruned structure to guarantee the alignment of the locations
of pruned filters across different layers. SR-LUT \cite{jo2021practical} uses look-up tables to  retrieve the precomputed HR output values for  LR input pixels, with a more significant SR performance degradation.
However, these SR models do not consider the actual mobile deployment, and the sizes of the models are still too large. The actual SR deployment is rarely investigated. 
The winner of the PIRM challenge \cite{vu2018fast}, MobiSR \cite{lee2019mobisr}, and work \cite{Zhan2021AchievingOR} explore the on-device SR, but the models take seconds for a single image, far from real time, or require long search time. Work \cite{ignatov2021real} considers real-time SR   deployed on the powerful mobile TPU, which is not  widely adopted such as mobile CPU/GPU.

\noindent \textbf{NAS for SR.} 
NAS has been shown to outperform heuristic networks in various applications. Recent SR works start to leverage NAS to find efficient, lightweight, and accurate SR models. Works \cite{chu2019fast,chu2020multi,Zhan2021AchievingOR} leverage   reinforced evolution algorithms to achieve SR as a multi-objective problem. Work~\cite{ahn2021neural} uses a hierarchical search strategy to find the connection with local and global features. LatticeNet \cite{luo2020latticenet} learns the combination of residual blocks with the attention mechanism. Work~\cite{wu2021trilevel,huang2021lightweight,ding2021hrnas} search  lightweight architectures  at  different levels with  differentiable architecture search (DARTS) \cite{liu2018darts}. 
DARTS based methods introduce architecture hyper-parameters which are usually continuous rather than binary, incurring additional bias during  selection and  optimization.
Furthermore, the above-mentioned methods typically take the number of parameters or MACs into the objective function, rather than  on-mobile latency as discussed in Sec.~\ref{sec: speed_incorporation}. Thus they can hardly satisfy the real-time requirement.

\noindent \textbf{Hardware Acceleration.} 
A significant emphasis on optimizing the DNN execution has emerged in recent years \cite{lane2016deepx,xu2018deepcache,huynh2017deepmon,yao2017deepsense,han2016mcdnn,niu2020patdnn,dong2020rtmobile,jian2021radio,gong2022automatic}. There are several representative DNN acceleration frameworks including Tensorflow-Lite \cite{TensorFlow-Lite}, Alibaba MNN \cite{Ali-MNN}, Pytorch-Mobile \cite{Pytorch-Mobile}, and TVM \cite{chen2018tvm}. These frameworks include several graph optimization techniques such as layer fusion, and constant folding.

\section{Motivation and Challenges} \label{sec:challenges}

With the rapid development of mobile devices and real-time applications such as live streaming, it is essential and desirable to implement real-time SR on resources-limited mobile devices. However, it is challenging. 
To maintain or upscale the spatial dimensions of feature maps based on large input/output  size,  SR  models  typically  consume  tens  of  or   hundreds of GMACs   (larger than several GMACs in image classification \cite{liu2019metapruning,wan2020fbnetv2}), incurring difficulties for real-time inference. For example, prior works on mobile SR deployment \cite{lee2019mobisr} and \cite{vu2018fast} achieve 2792ms and 912ms on-mobile inference latency, respectively, far from real-time.

We can adopt NAS or pruning methods to find a lightweight SR model with fast speed on mobile devices. But there are several challenges: 
(\texttt{C1}) tremendous searching overhead with NAS, (\texttt{C2}) misleading  magnitude during pruning,  (\texttt{C3}) speed incorporation issues, and (\texttt{C4}) heuristic depth determination.

\noindent \textbf{Tremendous Searching Overhead with NAS.} 
In NAS, the exponentially growing search space leads to tremendous search overhead. Specifically, the RL-based \cite{zoph2016neural,zhong2018practical,zoph2018learning} or evolution-based NAS methods \cite{real2019regularized,tan2019efficientnet,yang2020cars} typically need to sample large amounts of candidate models from the search space and train each candidate architecture with multiple epochs, incurring long search time and high computation cost.  
Besides,  differentiable NAS methods \cite{brock2017smash,bender2018understanding,liu2018darts} build super-nets to train multiple architectures simultaneously, causing significant memory cost and limited discrete search space up-bounded by the available memory.
To mitigate these, there are certain compromised strategies, such as proxy tasks (to search on CIFAR and target on ImageNet) \cite{real2019regularized,zhou2020econas,yang2020cars} and performance estimation (to predict/estimate the architecture performance with some metrics) \cite{abdelfattah2021zerocost,tanaka2020pruning,ming_zennas_iccv2021}.

\noindent \textbf{Misleading Magnitude during Pruning.} 
Pruning can also be adopted to reduce the model size, which  determines the per-layer pruning ratio and pruning positions. 
With the assumption that weights with smaller magnitudes are less important for final accuracy, magnitude-based pruning \cite{han2016deep_compression,mao2017exploring,he2017channel,zhang2021unified,wen2016learning,gong2020privacy,li2020ss,ma2020blk,yuan2021mest} is widely employed to prune weights smaller than a threshold. However, the assumption is not necessarily true, and weight magnitudes can be misleading. Magnitude-based pruning is not able to achieve importance shifting during pruning. As detailed in Appendix~\ref{app: small_magenitude}, in iterative magnitude pruning, small weights pruned first are not able to become large enough  to contribute to the accuracy. Thus layers pruned more at initial will be pruned more and more, causing a non-recoverable pruning policy. 
It becomes pure exploitation without exploration.  

\label{sec: speed_incorporation}
\noindent\textbf{Speed Incorporation Issues.}
\label{sec: speed_incorporation}
To achieve real-time inference on mobile, it is essential to obtain the on-mobile speed performance when searching  architectures. However, it is non-trivial to achieve this since 
testing speed requires an additional process to interact with the mobile device for a few minutes, which can hardly be incorporated into a typical model training.  
To mitigate this, certain methods \cite{liu2019metapruning,wan2020fbnetv2,9522982} adopt weight number or computation counts  as an estimation of the speed performance.   Other methods \cite{wu2019fbnet,dai2019chamnet,yang2018netadapt}  first collect on-mobile speed data and then build lookup tables with the speed data to estimate the speed. 

\noindent\textbf{Heuristic Depth Determination.}
Reducing model depth can avoid all computations in the removed layers, thus significantly accelerating the inference. Since previous NAS works do not incorporate a practical speed constraint or measurement during optimization,  their search on model depth is usually heuristic. Designers determine the model depth according to a simple rule that the model should satisfy an inference budget, without a specific optimization method \cite{ming_zennas_iccv2021,liu2018progressive,zhou2020econas,yang2020cars,bender2018understanding,liu2018darts}. 
More efforts are devoted to searching other optimization dimensions such as kernel size or width rather than model depth.

\section{Our Method}
\label{sec:method}

We first introduce the framework, then discuss the components of the framework in detail. We also specify how it can deal with the challenges   in Sec. \ref{sec:challenges}. 
\subsection{Framework with Adaptive SR Block}

\begin{figure}[t]
     \centering
     \includegraphics[width=0.78\columnwidth]{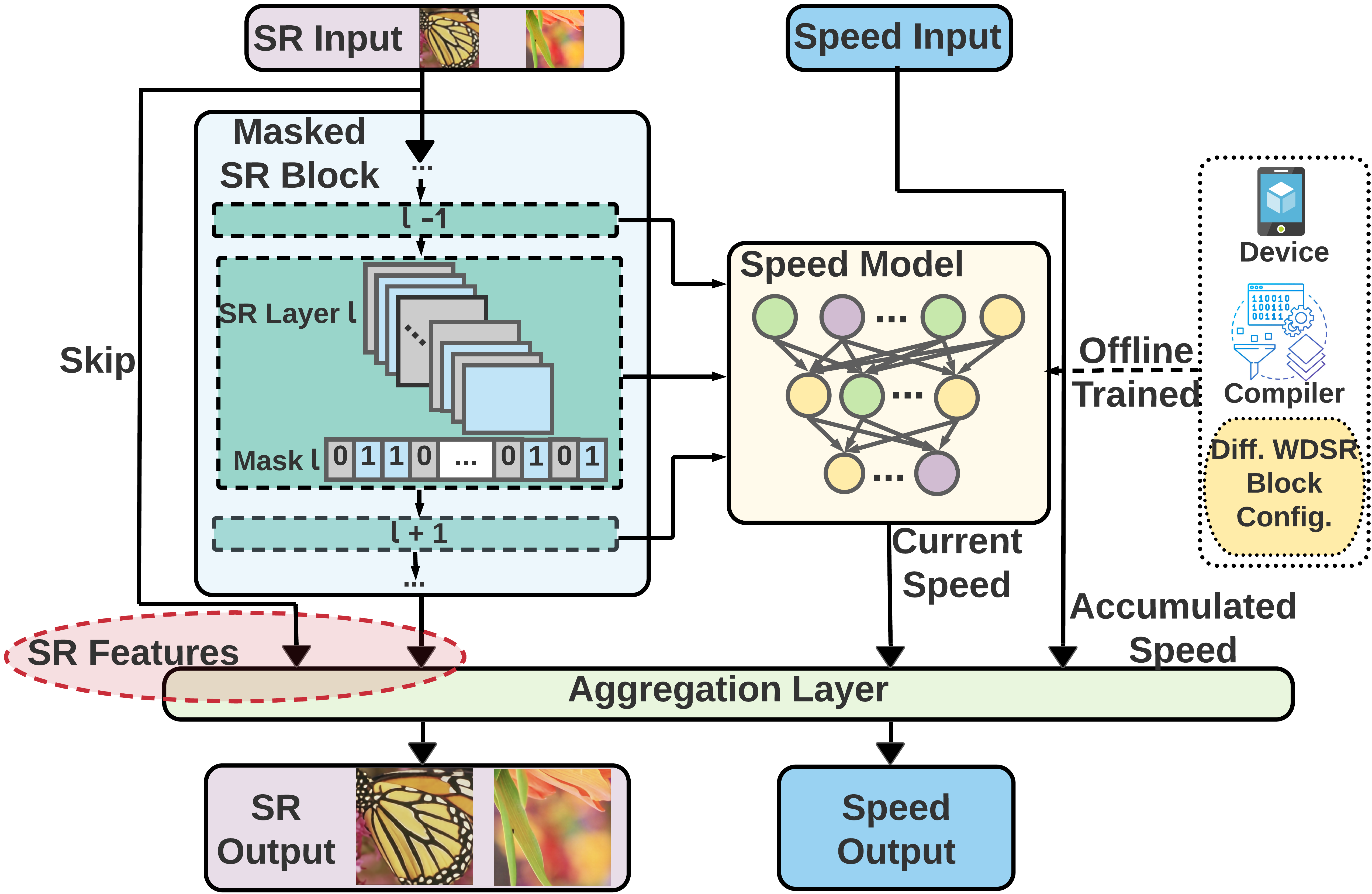}  
     \caption{Architecture of the adaptive SR block search.}
    \label{fig:search}  
\end{figure}

In the framework, we perform a compiler-aware architecture depth and per-layer width search to achieve real-time SR on mobile devices. The search space contains the width for each CONV layer and the number of stacked SR blocks in the model, which is too large to be explored with a heuristic method. Therefore, we propose an adaptive SR block to implement the depth and per-layer width search, and the model is composed of multiple adaptive SR blocks. Fig.~\ref{fig:search} shows the architecture of the adaptive SR block. It consists of a masked SR block, a speed model, and an aggregation layer. The adaptive SR block has two inputs (and outputs) corresponding to the SR features and the accumulated speed, respectively. It achieves per-layer width search with mask layers in the masked SR blocks and depth search with aggregation layer to choose a path between the skip connection and the masked SR block. Besides, to obtain the on-mobile speed performance, we adopt a speed model to predict the speed of the masked SR block. The speed model is trained on our own dataset with speed performance of various block width configurations measured through compiler optimizations for significant inference acceleration to achieve accurate speed prediction.  

\subsection{Per-Layer Width Search with Mask Layer for \texttt{C1} and \texttt{C2}}
Width search is performed for each CONV layer in a typical WDSR block \cite{yu2018wide}. WDSR is chosen as our basic building blocks since it has demonstrated high efficiency in SR tasks \cite{Yu_2019_ICCV,yu2019autoslim,Cheng_2019_CVPR_Workshops}. Note that our framework is not limited to the WDSR block and can be easily extended to various residual SR blocks \cite{ahn2018fast,lim2017enhanced,hui2018fast} in the literature.
To satisfy the real-time requirement, we perform a per-layer width search to automatically select an appropriate number of channels for each CONV layer in the  WDSR block. 
Specifically, we insert  a differentiable mask layer (a depth-wise $1\times 1$ CONV  layer) after each CONV layer to serve as the layer-wise trainable mask, as shown below, 
{\small
\begin{equation}
  \bm  a_l^n =  \bm m_l^n \odot (\bm w_l^n \odot \bm a_{l-1}^n),
\end{equation}}%
where $\odot$ denotes the convolution operation. $\bm w_l^n \in R^{o\times i\times k\times k}$ is the weight parameters in the $l^{th}$ CONV layer of the $n^{th}$ block, with $o$ output channels, $i$ input channels, and  kernels of size $k\times k $. 
$\bm a_l^n \in R^{B\times o\times s\times s'}$ represents the output features of $l^{th}$ layer (with the trainable mask), with $o$ channels and  $s\times s'$ feature size. $B$ denotes the batch size. 
$\bm m_l^n \in R^{o\times 1\times 1\times 1}$ is the corresponding weights of the depth-wise CONV layer (i.e., the mask layer).

We use each element of $\bm m_l^n$  as the pruning indicator for the corresponding output channel of  $\bm w_l^n \odot \bm a_{l-1}^n$.
Larger elements of $\bm m_l^n$  mean that the corresponding channels should be preserved while smaller elements indicate pruning the  channels. 
Formally, we use a threshold to convert    $\bm m_l^n$ into a binary mask,
{\small
\begin{equation}
 \bm b_l^n = 
    \begin{cases}
    1, \bm m_l^n > thres. \\
    0, \bm m_l^n \leq thres.
    \end{cases} \text{(element-wise)}, 
\end{equation}}%
where $\bm b_l^n \in \{0, 1\}^{o\times 1\times 1\times 1}$ is the binarized $\bm m_l^n$.
We initialize $\bm m_l^n$  with random values between 0 and 1, and the adjustable  $thres$ is  set to 0.5 in our case.  The WDSR block with the proposed mask layers   is named as masked SR block. 

Thus we are able to obtain a binary mask for each CONV layer. The next  problem  is how to make the mask trainable, as the binarization operation is  non-differentiable, leading to difficulties for back-propagation. To solve this,  we integrate Straight Through Estimator (STE) \cite{bengio2013setimating} as shown below, 
{\small
\begin{equation}
\frac{\partial \mathcal L}{\partial \bm m_l^n} = \frac{\partial \mathcal L}{\partial \bm b_l^n},
\end{equation}}%
where we directly pass the gradients through the binarization.  The STE method is originally proposed to avoid the non-differentiable problems in quantization tasks \cite{liu2019learning,yin2019understanding}.  
Without STE, some methods adopt complicated strategies to deal with the non-differentiable binary masks  such as \cite{guo2020dmcp,guan2020dais}.  

With the binarization and the STE method, we are able to build a trainable mask to indicate whether the corresponding channel is pruned or not. 
Our mask generation and training are more straightforward and simpler. For example,  proxyless-NAS \cite{cai2018proxylessnas} transforms the real-valued weights  to  binary gates  with  a  probability distribution, and adopts complex mask updating procedure (such as task factorizing). SMSR \cite{wang2021exploring} adopts Gumbel softmax to perform complex sparse mask CONV.  Unlike  proxylessNAS or SMSR,  we  generate binary  masks  simply  via  a  threshold  and  train  the  masks directly  via  STE.

 \subsection{Speed Prediction with Speed Model for \texttt{C3}}
 \label{sec: speed_model_framework}

To achieve real-time SR inference on mobile devices, we take the inference speed into the optimization to satisfy a given real-time threshold.  
It is hard to measure the practical speed or latency of various structures on mobile devices during optimization. Traditionally, the inference speed may be estimated roughly with the number of computations \cite{liu2019metapruning,wan2020fbnetv2,9522982} or a latency lookup table \cite{wu2019fbnet,dai2019chamnet,yang2018netadapt}, which can hardly provide an accurate speed. 
To solve this problem, we adopt a DNN-based speed model to predict the inference speed of the block. 
The input of the speed model is the width of each CONV layer in the block, and it outputs the block speed. As shown in Fig. \ref{fig:search},  the width of each CONV layer can be obtained through the mask layer. Thus the speed model can work perfectly with the width search,  dealing with \texttt{C3} to provide speed performance of various architectures. 

To train such a speed model, we first need to build a speed dataset with block latency  of various layer width configurations in the block. Next, we can train a speed model based on the dataset to predict the speed. We find that the trained speed model is accurate in predicting the speed of different layer widths in the block (with 5\% error at most). 
We show the details about the dataset, speed model, and the prediction accuracy in Sec.~\ref{sec: speed_model_compiler_optimization} and Appendix B.

We highlight that our speed model  not only takes the masks as inputs to predict the  speed, but also back-propagates the gradients from the speed loss  (Eq.~\eqref{eq:loss}) to update the masks as detailed in Sec.~\ref{sec: training_loss}, rather than just predicting performance forwardly such as 
\cite{wen2020neural}. That is why we  build the speed model based on DNNs instead of loop-up tables.  The trainable masks and the  speed model are combined comprehensively to solve the problem more efficiently.

\subsection{Depth Search with Aggregation Layer for \texttt{C4}}
Although reducing the per-layer width can  accelerate the inference, 
removing the whole block can avoid the computations of the whole block, thus providing higher speedup. Hence, besides width search, we further incorporate depth search to automatically determine the number of adaptive SR blocks in the model. Note that although per-layer width search may also converge to zero width, which eliminates the entire block, we find that in most cases, there are usually a few channels left in each block to promote the SR performance, leading to difficulties in removing the whole block. Thus it is necessary to incorporate depth search.

To perform depth search, we have two paths in each adaptive SR block. As shown in Fig.~\ref{fig:search}, one path is the skip connection, and the other path is the masked SR block. In the aggregation layer, there is a parameter like a switch to control which path the SR input goes through. If the SR input chooses the skip path, the masked SR block is skipped, and the latency of this block is just 0, leading to significant inference acceleration. 
The aggregation layer plays a key role in the path selection. It contains two trainable parameters $\alpha_s$ and $\alpha_b$. In the forward pass, it selects the skip path or the masked WDSR block path based on the relative relationship of $\alpha_s$ and $\alpha_b$, as shown below,
{\small
\begin{align}
  \bm  \beta_s = 0 \   \text{and}  \ \beta_b =1,  \text{if} \ \alpha_s \le \alpha_b,  \\  
    \bm  \beta_s = 1 \  \text{and} \ \beta_b =0,   \text{if} \  \alpha_s > \alpha_b,  
\end{align}}%
where the binarized  variables $\beta_s$ and $\beta_b$   denote the path selection ($\beta_s {=}1$ means choosing the skip path and $\beta_b{=}1$ means choosing the masked SR block path).
Since the comparison operation is non-differentiable, leading to difficulties for back-propagation, similarly we adopt STE  \cite{bengio2013setimating} to make it differentiable as below,
{\small
\begin{align}
\frac{\partial \mathcal L}{\partial  \alpha_s} = \frac{\partial \mathcal L}{\partial \beta_s},  \  \ \frac{\partial \mathcal L}{\partial  \alpha_b} = \frac{\partial \mathcal L}{\partial \beta_b}.
\end{align}}

In the aggregation layer, 
the forward computation can be represented below,  
{\small
\begin{align}
\bm {a}^n & = \beta_s \cdot \bm a^{n-1}  + \beta_b \cdot \bm a_L^n, \\
v_n & = v_{n-1} + \beta_b \cdot  v_c, 
\end{align}}
where $\bm {a}^n$ is the SR output features of the $n^{th}$ adaptive SR block.   $\bm {a}^n_L$ is the SR output features of masked SR block in the $n^{th}$ adaptive SR block, and $L$ is the maximum number of CONV layers in each block and we have $l{\le} L$.
$v_n$ is the accumulated speed or latency until the $n^{th}$ adaptive SR block and  $v_c$ is the speed of the current block which is predicted by the speed model. 
By training  $\alpha_s$ and $\alpha_b$, the model can learn to  switch between the skip path and the  SR  path to determine the model depth, thus dealing with \texttt{C4}.

\subsection{Training Loss} \label{sec: training_loss}

Multiple adaptive SR blocks can form the SR model, which provides two outputs including the typical SR outputs and the speed outputs. The training loss is a combination of a typical SR loss and a speed loss as below,
{\small
\begin{align}
\mathcal L_{SPD} &= \text{max}\{0, v_N- v_T \}, \\
\mathcal L & =  \mathcal L_{SR} + \gamma \mathcal L_{SPD}, \label{eq:loss}
\end{align}}%
where $v_T$ is the real-time threshold, $v_N$ is the accumulated speed of $N$ blocks, and
$\gamma$ is a parameter to control their relative importance. The objective is to achieve high SR performance while the speed can satisfy a real-time threshold.
To summarize, with the trainable masks, 
the speed model, and the aggregation layer in the adaptive SR block, our search algorithm achieves the following advantages: 
\begin{itemize}[leftmargin=*]
    \item The mask can be trained along with the  network parameters via  gradient descent optimizers, thus dealing with \texttt{C1} to save search overhead compared with previous one-shot pruning \cite{he2017channel,frankle2018lottery} or NAS methods \cite{zoph2016neural,zhong2018practical} to train multiple epochs for each candidate architecture with huge searching efforts. 
    \item Compared with magnitude-based threshold pruning, we decouple the trainable masks from original model parameters, thus enabling exploitation and overcoming the drawbacks of magnitude-based  pruning,  dealing with \texttt{C2}.  
    \item We use the speed model for predicting the speed to solve \texttt{C3}, which is differentiable regarding the trainable mask. Thus the mask is trained to find a model with both high SR performance and fast inference speed. 
    \item We also incorporate depth search though  aggregation layers  to deal with \texttt{C4}.
\end{itemize}

\section{Compiler Awareness with Speed Model } 
\label{sec: speed_model_compiler_optimization}

To satisfy the speed requirement with a given  latency threshold on a specific mobile device,  it is required to obtain the actual inference latency   on the  device. 
It is non-trivial to achieve this as the model speed varies with different model width and depth. 
It is unrealistic to measure the actual on-mobile speed   during the   search, as the search space is quite large,  
and testing the mobile speed of each candidate can take a few minutes, which is not compatible with  DNN training. 

To solve this problem,  
we adopt a  speed model to predict the inference latency of the masked SR block with various width configurations. With the speed model, we can obtain the speed prediction as outputs by providing the width of each CONV layer in the SR block as inputs. It is fully compatible with the trainable mask, enabling differentiable  model speed with respect to the layer width.

To obtain the speed model, we first build a latency dataset with latency data measured on the hardware platforms incorporated with compiler optimizations. Then the DNN speed model is trained based on the latency dataset. 

\begin{figure}[t]
     \centering
     \includegraphics[width=0.8\columnwidth]{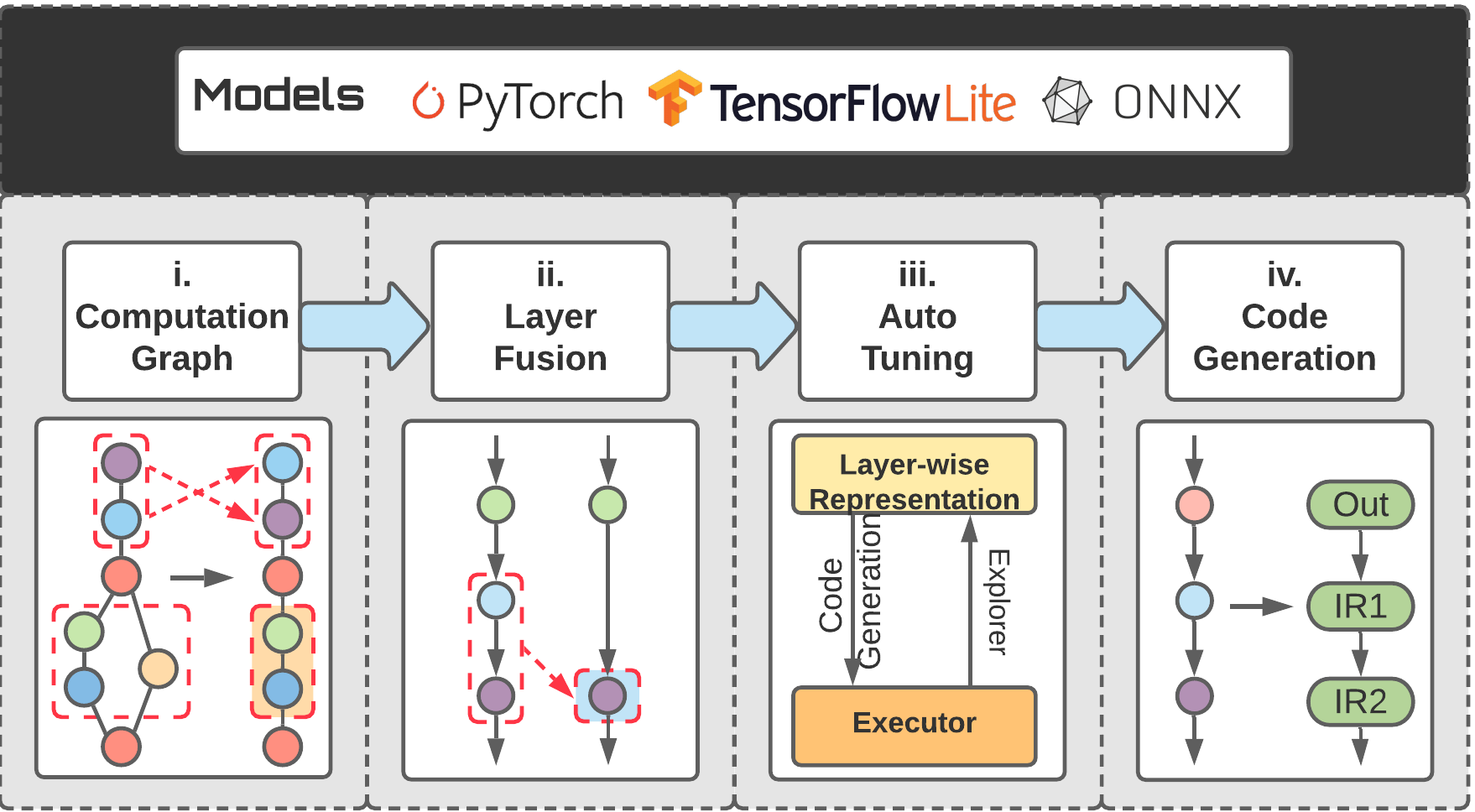} 
     \caption{The overview of compiler optimizations.}
    \label{fig:compiler}  
\end{figure}

\noindent\textbf{Compiler Optimization.}
To build a latency dataset, we need to measure the speed of various block configurations on mobile devices. Compiler optimizations are adopted to accelerate the inference speed during speed testing. It is essential to incorporate compiler optimizations as they can significantly accelerate the inference speed. 
The overview of the compiler optimizations is shown in Fig. \ref{fig:compiler}.

To fully exploit the parallelism for a higher speedup,  the key features of SR have to be considered. As the objective of SR is to obtain a HR image from its LR counterpart, each layer has to  maintain or upscale the spatial dimensions of the feature, leading to larger feature map size and more channels compared with classification tasks. Therefore, the data movements between the memory and cache are extremely intensive. To reduce the data movements for faster inference, we adopt two important optimization techniques: 1) operator fusion and 2) decreasing the amount of data to be copied between CPU and GPU. 

Operator fusion is a key optimization technique adopted in many state-of-the-art DNN execution framework \cite{TensorFlow-Lite,Ali-MNN,Pytorch-Mobile}. However, these frameworks usually adopt fusion approaches based on certain patterns that are too restrictive to cover the diversity of operators and layer connections. To address this problem, we classify the existing operations in the SR model into several groups based on the mapping between the input and output, and develop rules for different combinations of the 
groups in a more aggressive fusion manner. For instance, CONV operation and depth-to-space operation can be fused together.  With layer fusion, both the memory consumption of the intermediate results and the number of operators can be reduced.  An auto-tuning process is followed to determine the best-suited configurations of parameters for different mobile CPUs/GPUs and Domain Specific Language (DSL) based code generation. After that, a high-level DSL is leveraged to specify the operator in the computational graph of a DNN model. We show more details about compiler optimization in Appendix C.

\noindent\textbf{Latency Dataset.}
To train the speed model, we first  measure and collect the inference speed of the WDSR block under various CONV layer width configurations. After that, a dataset of the WDSR block on-mobile speed with different configurations can be built.  We vary the number of filters in each CONV layer as the different width configurations. The inference time is measured on the target device (Samsung Galaxy S21) by stacking 20  WDSR blocks with the same configuration, and the average latency is 
used as the inference time  to mitigate the overhead of loading data on mobile GPU. As the maximum number of CONV layers in each masked WDSR block is $L$, each data point in the dataset can be represented as a tuple with $L{+}2$ elements: $\{\mathcal{F}_{CONV^1}, \cdots, \mathcal{F}_{CONV^{L+1}}, \mathcal{T}_{inference}\}$, where $\mathcal{F}_{CONV^i},$ for $i{\in}\{1,\cdots, L\}$, indicates the number of input channels for the $i^{th}$ CONV layer, $\mathcal{F}_{CONV^{L+1}}$ is the number of output channels for the last CONV layer, and $\mathcal{T}_{inference}$ is the inference speed for this configuration measured in milliseconds. The entire dataset is composed of 2048 data points.

\noindent\textbf{Speed Model.}
With the latency dataset, the speed model can be trained on the collected data points. The inference speed estimation  is a regression problem, thus, a  network with 6 fully-connected layers combined with ReLU activation is used as the speed model.  During the speed model training, 90\% of the data is used for training  and the rest is for  validation. After training, the speed model can predict the inference time of various   block configurations with high accuracy. From our results, the speed model only incurs 5\% of deviation for the speed prediction.
The speed model has two advantages: (1) It is compatible with the width search framework as the trainable mask can be directly fed into the speed model. (2) It makes  the   model speed    differentiable with respect to the  masks, and back-propagates   gradients to update the masks, thus the model can  update the model speed by adjusting the layer width though back-propagation.   

\section{Experiments}
\label{sec:exp}

\begin{table*}[t]
\scriptsize\sffamily
    \centering
    \renewcommand{\arraystretch}{1.2}
\begin{adjustbox}{max width=0.9\textwidth}
\begin{tabular}{c|c|c|c|c|cccc|cccc}
\toprule
\multirow{2}{*}{Scale} & \multicolumn{1}{c|}{\multirow{2}{*}{Method}} & \multicolumn{1}{c|}{\multirow{2}{*}{ \makecell[c]{ Params \\ (K)} }} & \multicolumn{1}{c|}{\multirow{2}{*}{\makecell[c]{  MACs \\(G)}} } & \multicolumn{1}{c|}{\multirow{2}{*}{ \makecell[c]{Latency \\ (ms)}}} & \multicolumn{4}{c|}{PSNR}                                                                                          & \multicolumn{4}{c}{SSIM}                                                                                          \\ \cline{6-13} 
                       & \multicolumn{1}{c|}{}                        & \multicolumn{1}{c|}{}                                  & \multicolumn{1}{c|}{}                      & \multicolumn{1}{c|}{}                              & \multicolumn{1}{c}{Set5} & \multicolumn{1}{c}{Set14} & \multicolumn{1}{c}{B100} & \multicolumn{1}{c|}{Urban100} & \multicolumn{1}{c}{Set5} & \multicolumn{1}{c}{Set14} & \multicolumn{1}{c}{B100} & \multicolumn{1}{c}{Urban100} \\ \hline \hline
\multirow{14}{*}{\footnotesize $\times$2}

    & \textsc{FSRCNN}~\cite{dong2016accelerating} & 12 & 6.0 & 128.47 & 37.00  & 32.63 & 31.53  & 29.88   & 0.9558 &   0.9088 &  0.8920 & 0.9020 \\
    & \textsc{MOREMNAS-C}~\cite{chu2020multi} & 25 & 5.5 & --- & 37.06  & 32.75 &   31.50 & 29.92 & 0.9561 & 0.9094 & 0.8904 &  0.9023 \\
    & \textsc{TPSR-NOGAN}~\cite{lee2020journey} & 60 &14.0 & --- &37.38  & 33.00 & 31.75 & 30.61  &  0.9583 &  0.9123 &  0.8942 & 0.9119\\
    & \textsc{LapSRN}~\cite{lai2017deep} & 813& 29.9& --- &  37.52 &  33.08 &31.80   &  30.41 & 0.9590 &0.9130 &0.8950 &0.9100 \\
    & \textsc{CARN-M}~\cite{ahn2018fast} & 412 &91.2& 1049.92 & 37.53 &  33.26 & 31.92 &  31.23 & 0.9583 &0.9141 &0.8960 &0.9193\\
    & \textsc{FALSR-C}~\cite{chu2019fast} & 408 &93.7 & --- & 37.66  & 33.26 & 31.96  &  31.24 & 0.9586 &0.9140 &0.8965 &0.9187 \\
    & {\textsc{ESRN-V}}~\cite{song2020efficient} & 324 & 73.4 & --- & 37.85 & 33.42 &32.10  &  31.79   & 0.9600 & 0.9161 & 0.8987 & 0.9248 \\
    & \textsc{EDSR}~\cite{lim2017enhanced} & 1518 & 458.0 & 2031.65 & 37.99  & 33.57 &  32.16 &  31.98 & 0.9604 & 0.9175 & 0.8994 & 0.9272 \\
    & \textsc{WDSR}~\cite{yu2018wide} & 1203 & 274.1 & 1973.31 & 38.10 &  33.72 &32.25  & 32.37  &  0.9608 & 0.9182 & 0.9004 & 0.9302 \\
    & { \textsc{SMSR}~\cite{wang2021exploring}}  & 985 & 131.6 & --- & 38.00 & 33.64 & 32.17 & 32.19  &   0.9601 & 0.9179 &  0.8990 & 0.9284 \\  
    & { \textsc{SRPN-L}~\cite{zhang2021learning}}  & 609 & 139.9& --- & 38.10 & 33.70 & 32.25 & 32.26  &   0.9608 & 0.9189 &  0.9005 & 0.9294 \\ 
    & \textbf{Ours ($v_T{=}100$}ms\textbf{)}  & 47 & 11.0 & \textbf{98.90} & 37.64  & 33.16 &  31.91 &  31.08 &  0.9591 & 0.9136 & 0.8961 & 0.9170 \\
    & \textbf{Ours ($v_T{=}70$}ms\textbf{)} & 28 & 6.6 & \textbf{66.09} & 37.49  & 33.05 & 31.81 & 30.76  &  0.9584 & 0.9123 & 0.8946 & 0.9135 \\
    & \textbf{Ours ($v_T{=}40$}ms, \textbf{real-time)}  & 11 & 2.5 & \textbf{34.92} & 37.19  &  32.80&  31.60 & 30.15  &0.9572 & 0.9099 &  0.8919 & 0.9054 \\
\hline
\multirow{16}{*}{\footnotesize $\times$4}     
    & \textsc{FSRCNN}~\cite{dong2016accelerating} & 12 & 4.6 & 98.13 & 30.71  &  27.59 & 26.98 & 24.62 & 0.8657& 0.7535 &0.7150 &0.7280 \\
    & {\textsc{TPSR-NOGAN}}~\cite{lee2020journey} & 61 &3.6 & 55.82 &31.10  &  27.95 &  27.15 & 24.97 & 0.8779 &0.7663 &0.7214 &0.7456 \\
    & {\textsc{FEQE-P}}~\cite{vu2018fast} & 96 & 5.6 & 82.81 & 31.53 & 28.21 &27.32  &25.32  & 0.8824 & 0.7714 & 0.7273 & 0.7583 \\
    & \textsc{CARN-M}~\cite{ahn2018fast} & 412 &32.5 & 374.15 & 31.92  & 28.42  &27.44  & 25.62 & 0.8903 &0.7762 &0.7304 &0.7694\\
    & {\textsc{ESRN-V}}~\cite{song2020efficient} & 324 & 20.7 & --- & 31.99  & 28.49  & 27.50 & 25.87 & 0.8919 &0.7779 &0.7331 &0.7782 \\
    & {\textsc{IDN}}~\cite{hui2018fast} & 600 & 32.3 & --- & 31.99  & 28.52  & 27.52 & 25.92 &  0.8928 &0.7794 &0.7339 &0.7801 \\
    & \textsc{EDSR}~\cite{lim2017enhanced} & 1518 & 114.5 & 495.90 & 32.09 & 28.58 & 27.57 & 26.04 & 0.8938 & 0.7813 & 0.7357 & 0.7849 \\
    &\textsc{DHP-20}~\cite{li2020dhp}& 790 & 34.1 & --- & 31.94  &28.42  & 27.47 &  25.69 & \quad--- & \quad--- & \quad--- &\quad--- \\ 
    &\textsc{IMDN}~\cite{hui2019lightweight}& 715 & 40.9 & --- & 32.21  &  28.58 &   27.56 & 26.04 & 0.8948 &0.7811 &0.7353 & 0.7838\\ 
    & {\textsc{WDSR}}~\cite{yu2018wide} & 1203 & 69.3 & 533.02 & 32.27 & 28.67  &  27.64 &26.26  & 0.8963 &0.7838 &0.7383 &0.7911 \\
    & {\textsc{SR-LUT-S}~\cite{jo2021practical} } &  77 & --- & --- & 29.77 &  26.99 & 26.57  & 23.94  &   0.8429 & 0.7372 &  0.6990 & 0.6971 \\  
    & {\textsc{SMSR}~\cite{wang2021exploring} } &  1006 & 41.6 & --- & 32.12 &  28.55 & 27.55  & 26.11  &   0.8932 & 0.7808 &  0.7351 & 0.7868 \\  
    & { \textsc{SRPN-L}~\cite{zhang2021learning}}  & 623 & 35.8& --- & 32.24 & 28.69 & 27.63 & 26.16  &   0.8958 & 0.7836 &  0.7373 & 0.7875 \\ 
    
    & \textbf{Ours ($v_T{=}100$}ms\textbf{)} & 188 & 10.8 & \textbf{93.50} &  32.02 &  28.50 & 27.51 & 25.83 & 0.8922 & 0.7778 & 0.7328 & 0.7769 \\

    & \textbf{Ours ($v_T{=}70$}ms\textbf{)} & 116 & 6.7 & \textbf{64.95} & 31.88  &  28.43 & 27.46 &25.69  & 0.8905 & 0.7760 & 0.7312 & 0.7715 \\
    
    & \textbf{Ours ($v_T{=}40$}ms,\textbf{real-time)} & 66 & 3.7 & \textbf{36.46} & 31.73  & 28.28  & 27.34  &25.44  & 0.8878 & 0.7725 &0.7281 & 0.7620 \\
     \bottomrule
    \multicolumn{13}{l}{$*$ Some latency results are not reported as the models are not open-source or contain operators that cannot run on mobile GPU. } \\
    \multicolumn{13}{l}{$\dag$ The latency results are measured on the GPU of Samsung Galaxy S21.  }
    
\end{tabular}

\end{adjustbox}
    \caption{Comparison  with SOTA efficient SR models for implementing 720p.} 
        \label{table:result_sr}
\end{table*}

\subsection{Experimental Settings}
\noindent\textbf{SR Datasets.} 
All SR models are trained on the training set of  DIV2K \cite{Agustsson_2017_CVPR_Workshops}  with 800 training images. For  evaluation, four benchmark datasets Set5 \cite{bevilacqua2012low}, Set14 \cite{yang2010image}, B100 \cite{martin2001database}, and Urban100 \cite{huang2015single} are used for test. The PSNR and SSIM are calculated on the luminance channel (a.k.a. Y channel) in the YCbCr color space. 

\noindent\textbf{Evaluation Platforms and Running Configurations.} The training codes are implemented with  PyTorch. 8 GPUs are used to conduct the search, which usually finishes in 10 hours. The latency is measured on the GPU of an off-the-shelf Samsung Galaxy S21 smartphone, which has the Qualcomm Snapdragon 888 mobile platform with a Qualcomm Kryo 680 Octa-core CPU and a Qualcomm Adreno 660 GPU. Each test takes 50 runs on different inputs with 8 threads on CPU, and all pipelines on GPU. The average time is reported.

\noindent\textbf{Training Details.}
$48\times48$ RGB image patches are randomly sampled from LR images for each input minibatch.
We use the architecture of WDSR with 16 blocks as the backbone  of our NAS process. 
Considering the huge input size of SR (normally nHD--$640{\times}360$ inputs or higher resolution for $\times2$ task), 
a compact version of the WDSR block is chosen to fit the mobile GPU, where the largest filer number for each  CONV layer is 32, 146, and 28, respectively.
The backbone is initialized with the parameters of the pretrained WDSR model. 
Traditional MAE loss is used to measure the differences between the SR image and the ground-truth as the SR loss. The parameter $\gamma$ in the training loss denoted as Eq.~\eqref{eq:loss} is set to 0.01. The first 20 epochs are used for the NAS process, and the following 30 epochs for fine-tuning the searched model. ADAM optimizers with $\beta_{1}{=}0.9$, $\beta_{2}{=}0.999$, and $\epsilon{=}\num{1e-8}$ are used for both model optimization and fine-tuning process. The learning rate is initialized as $\num{1e-4}$ and reduced by half at 10, 16 epochs and at 20, 25 epochs in the NAS and fine-tuning process, respectively. The details of the searched architecture are in Appendix D.

\noindent\textbf{Baseline Methods. } We compare with some traditional human-designed  SR models such as FSRCNN  and EDSR. Besides, some baselines optimizing the speed or hardware with NAS approaches are also covered. For example, TPSR-NOGAN, FALSR-C, ESRN-V optimize the SR efficiency to facilitate the deployment on end devices.   Moreover, we compare with some methods exploring the sparsity in SR models such as DHP, SMSR, and SRPN-L for efficient inference.

\begin{figure*}[t]
  \centering
    \includegraphics[width=0.9\textwidth]{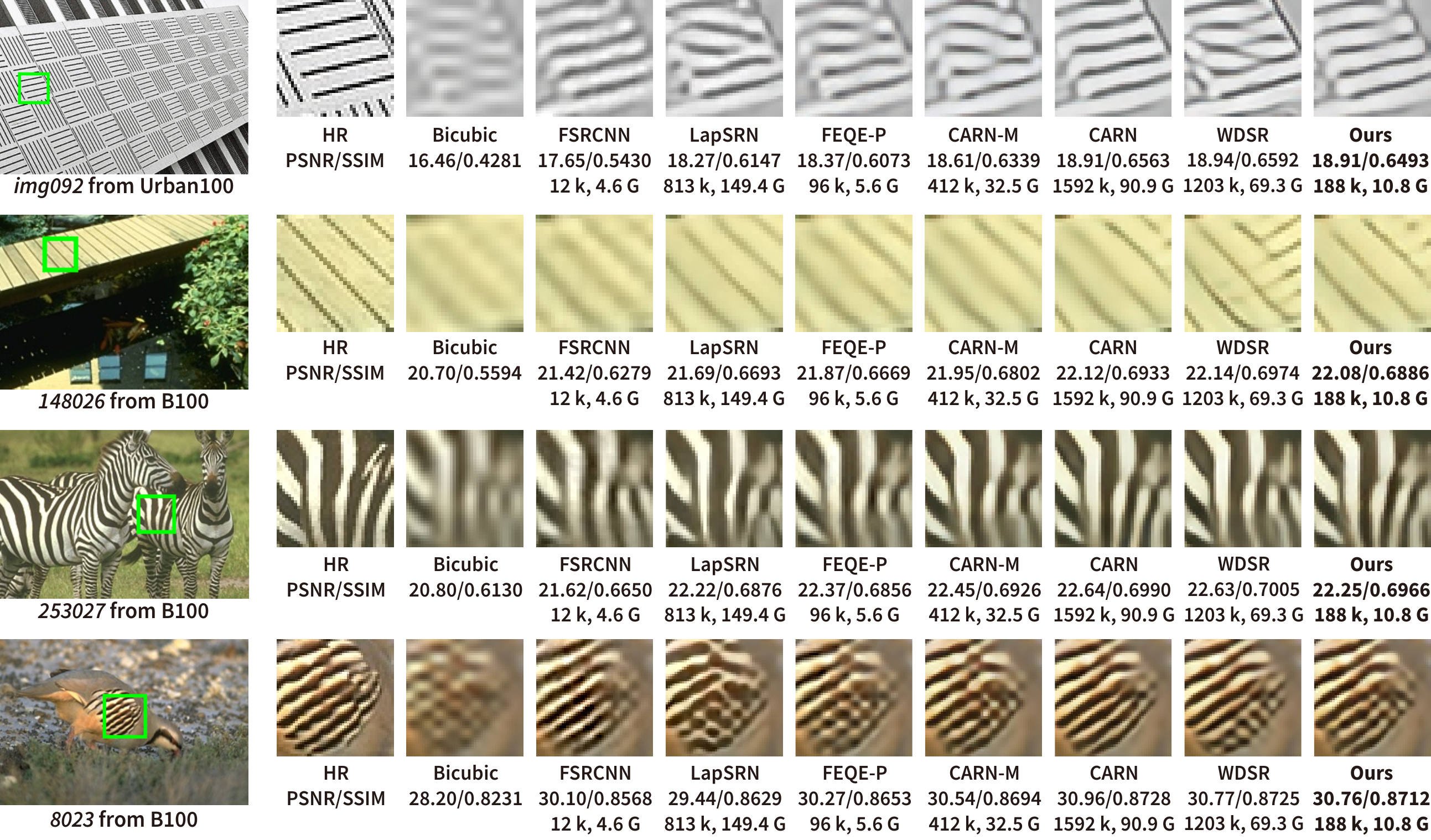}  
   \caption{Visual Comparisons with other   methods on Urban100/B100 for $\times$4 SR.}
   \label{fig:patch}
  \vspace{-0.2cm}
\end{figure*}

\subsection{Experimental Results}

\noindent\textbf{Comparison with Baselines on SR Performance.}
The comparisons of the models obtained by the proposed framework with state-of-the-art efficient SR works are shown in Table \ref{table:result_sr}. Two commonly used metrics (PSNR and SSIM) are adopted to evaluate image quality. The evaluations are conducted on $\times$2 and $\times$4 scales. For a fair comparison, we start from different low-resolution inputs but the high-resolution outputs are 720p ($1280{\times}  720$). To make a comprehensive study, the latency threshold  $v_T$ is set to different values. Specifically, as real-time execution typically requires at least 25 frames$/$sec (FPS), the latency threshold $v_T$ is set as 40ms to obtain SR models for real-time inference.

For $\times$2 scale, the model obtained with latency threshold $v_T$=100ms outperforms TPSR-NOGAN, LAPSRN, and CARN-M in terms of PSNR and SSIM with fewer parameters and MACs. Compared with FALSR-C, ESRN-V, EDSR, WDSR, SMSR, and SRON-L, our model greatly reduces the model size and computations with a 
competitive image quality performance. By setting $v_T$ as 70ms, our model has similar parameters and MACs as MOREMNAS-C, but achieves higher PSNR and SSIM performance.  Similar results can be obtained on the $\times$4 scale. 
Furthermore, for both scales, by setting $v_T$ as 40ms, we obtain extremely lightweight models and the models still maintain satisfying PSNR and SSIM performance on all   four datasets. Although SR-LUT  uses look-up tables for efficient SR inference, it suffers from more significant SR performance degradation.

The visual comparisons with  other SR methods for $\times$4 up-scaling task are shown in Fig.~\ref{fig:patch}. Our model can recover the details comparable or even better than other methods by using fewer parameters and computations.

\noindent\textbf{Comparison with Baselines on Speed Performance.} 
In general,  our method can achieve real-time SR inference (higher than 25 FPS) for implementing 720p resolution up-scaling with competitive image quality in terms of PSNR and SSIM on mobile platforms (Samsung Galaxy S21). Compared with \cite{wang2021exploring} which also explore the sparsity of SR models, our method can achieve more significant model size and computation reduction (our 11GMACs v.s. 131.6GFLOPs \cite{wang2021exploring}  for $\times$2 scale), leading to faster speed (our 11.3ms v.s. 52ms \cite{wang2021exploring} on Nvidia A100 GPU).

\begin{figure}[t]
\begin{floatrow}
\ffigbox{
  \includegraphics[width=0.9\linewidth]{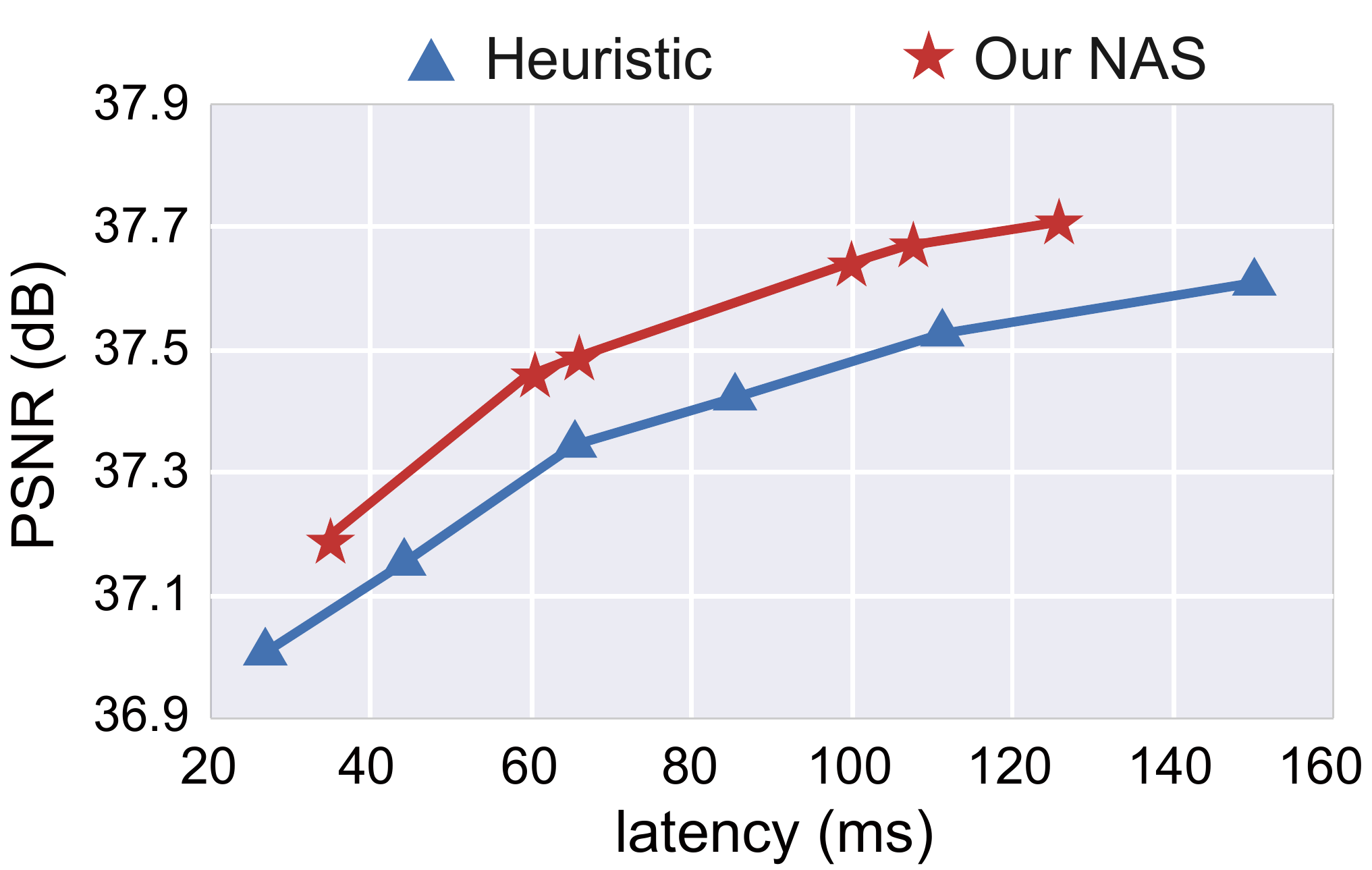}
}{
  \caption{Comparison of $\times2$ SR results between searched models and heuristic models on Set5  with latency measured on the GPU of Samsung Galaxy S21.}
  \label{fig:search_vs_human}
}

\capbtabbox{
\scriptsize\sffamily
    \centering
  \renewcommand{\arraystretch}{1.1}
\begin{adjustbox}{max width=0.45\textwidth}
\begin{tabular}{cc|c|cc|cc}
\toprule
\multicolumn{2}{c|}{Search Method}                & \multirow{2}{*}{ \makecell[c]{Latency \\ (ms)}} & \multicolumn{2}{c|}{Set 5 }    & \multicolumn{2}{c}{Urban100 } \\ \cline{1-2} \cline{4-7} 
\multicolumn{1}{c|}{ \makecell[c]{Width \\ Search} } & \makecell[c]{Depth\\ Search} &                               &  PSNR  & SSIM &  PSNR  & SSIM \\ 
 \hline   \hline
\multicolumn{1}{c|}{\xmark }            &       \xmark      &   150.92 & 37.62 & 0.9589     & 31.03 & 0.9164    \\  \hline
\multicolumn{1}{c|}{\xmark }            &       \cmark      &   111.58 & 37.65 & 0.9591     & 31.10	& 0.9172    \\  \hline
\multicolumn{1}{c|}{\cmark }            &       \xmark      &   108.38 & 37.65 & 0.9591     & 31.02	& 0.9161    \\  \hline  
\multicolumn{1}{c|}{\cmark }            &       \cmark      &   \textbf{98.90}  & 37.64 & 0.9591     & 31.08	& 0.9170    \\  
\bottomrule
\end{tabular}
\end{adjustbox}
}{
    \caption{Comparison of different search schema for $\times2$ scales. The performance is  evaluated on Set5 and Urban100 datasets}
    \label{table:ablation}
}
\end{floatrow}
\end{figure}

\noindent\textbf{Comparison with Heuristic Models.}
We  compare  our searched models with heuristic models, which are obtained by evenly reducing the depth and width from the WDSR model. Since we do not search per-layer width in heuristic models, the width is the same among all blocks in one heuristic model. For a fair comparison, the same compiler optimization framework is adopted for both searched models and heuristic models.  As shown in Fig.~\ref{fig:search_vs_human}, we can see that the NAS approach can achieve faster inference than the heuristic models under the same PSNR, demonstrating the effectiveness of the search approach.

\noindent\textbf{Compiler Optimization Performance.}
To demonstrate the effectiveness of our compiler optimizations, we implement CARN-M \cite{ahn2018fast}, FSRCNN \cite{dong2016accelerating}, and our searched model with the open-source MNN framework. 
By comparing  their PSNR and FPS performance, we find that our model can achieve higher FPS and PSNR  than the baseline models, with detailed results  in Appendix E.  We also compare with the compilation of \cite{huynh2017deepmon}  detailed in Appendix F.

\noindent\textbf{Performance on Various Devices.}
Our main results are trained and tested on the mobile GPU. We highlight that our method can be easily applied to all kinds of devices with their corresponding speed models. To demonstrate this, we perform compiler optimizations for the DSP on the mobile device and train the corresponding speed model. With the new speed model, we   use our method to search an SR model for the DSP, which can achieve 37.34 PSNR on   Set5   with 32.51 ms inference speed for $\times2$ up-scaling task, detailed  in Appendix G.

\subsection{Ablation Study}
For the ablation study, we investigate the influence of depth search and per-layer width search separately for $\times$2 scale task. 
Multiple runs are taken for each search method with different latency threshold $v_T$ so that the searched models have similar PSNR and SSIM on Set5 to provide a clear comparison. From the results in Table \ref{table:ablation}, we can see that both depth search only and width search only can greatly reduce the latency with better image quality than non-search case. Specifically, as a missing piece in many prior SR NAS works, depth search provides better PSNR and SSIM performance than width search 
on Urban100 with a slightly higher latency, which shows the importance of this search dimension. By combining depth search and width search, we could reach faster inference with similar PSNR and SSIM than conducting either search alone.

\section{Conclusion} \label{sec:conclusion}
We propose a compiler-aware NAS framework to achieve real-time SR on mobile devices. An adaptive WDSR block is introduced to conduct depth search and per-layer width search. The latency is directly taken into the optimization objective with the leverage of a speed model incorporated with compiler optimizations. 
With the framework, we achieve real-time SR inference for the implementation of 720p with competitive SR performance on mobile. \\

\noindent\textbf{Acknowledgments.}
The research reported here was funded in whole or in part by the Army Research Office/Army Research Laboratory via grant W911-NF-20-1-0167 to Northeastern University. Any errors and opinions are not those of the Army Research Office or Department of Defense and are attributable solely to the author(s). This research is also partially supported by National Science Foundation CCF-1937500 and CNS-1909172.

\clearpage
%
%
\bibliographystyle{splncs04}
\bibliography{egbib}

\end{document}